\documentclass[10pt,twocolumn,letterpaper]{article} 

\usepackage{avss}
\usepackage{times}
\usepackage{epsfig}
\usepackage{graphicx}
\usepackage{amsmath}
\usepackage{amssymb}
\usepackage{pifont}
\newcommand{\cmark}{\ding{51}}%
\newcommand{\xmark}{\ding{55}}%
\usepackage{fancyhdr}
\fancypagestyle{specialfooter}{%
  \fancyhf{}
  
  \fancyfoot[L]{978-1-5386-9294-3/18/\$31.00 ©2019 IEEE}
}

\avssfinalcopy 

\def\avssPaperID{18} 

\ifavssfinal\fi
\begin{document}

\title{Future Frame Prediction Using Convolutional VRNN for Anomaly Detection}

\author{Yiwei Lu, Mahesh Kumar Krishna Reddy, Seyed shahabeddin Nabavi and Yang Wang\\
University of Manitoba, Winnipeg, MB, Canada\\
{\tt\small \{luy2,kumarkm,nabaviss,ywang\}@cs.umanitoba.ca}
}
\maketitle
\thispagestyle{empty}

\thispagestyle{specialfooter}
\begin{abstract}
Anomaly detection in videos aims at reporting anything that does not conform the normal behaviour or distribution. However, due to the sparsity of abnormal video clips in real life, collecting annotated data for supervised learning is exceptionally cumbersome. Inspired by the practicability of generative models for semi-supervised learning, we propose a novel sequential generative model based on variational autoencoder (VAE) for future frame prediction with convolutional LSTM (ConvLSTM). To the best of our knowledge, this is the first work that considers temporal information in future frame prediction based anomaly detection framework from the model perspective.
 Our experiments demonstrate that our approach is superior to the state-of-the-art methods on three benchmark datasets.
\end{abstract}

\section{Introduction}
Anomaly detection is an essential problem in video surveillance. Due to the massive amount of available video data from surveillance cameras, it is time-consuming and inefficient to have human observers watching surveillance videos and report any anomalies. Ideally, we want an automatic system that can report abnormal events. Anomaly detection is challenging since the definition of ``anomaly'' is broad and ambiguous -- anything that deviates expected behaviours can be considered as ``anomaly''. It is infeasible to collect labeled training data that cover all possible anomalies. As a result, recent work in anomaly detection has focused on unsupervised approaches that do not require human labels.

\begin{figure}
    \centering
    \includegraphics[height=6cm, width=8cm]{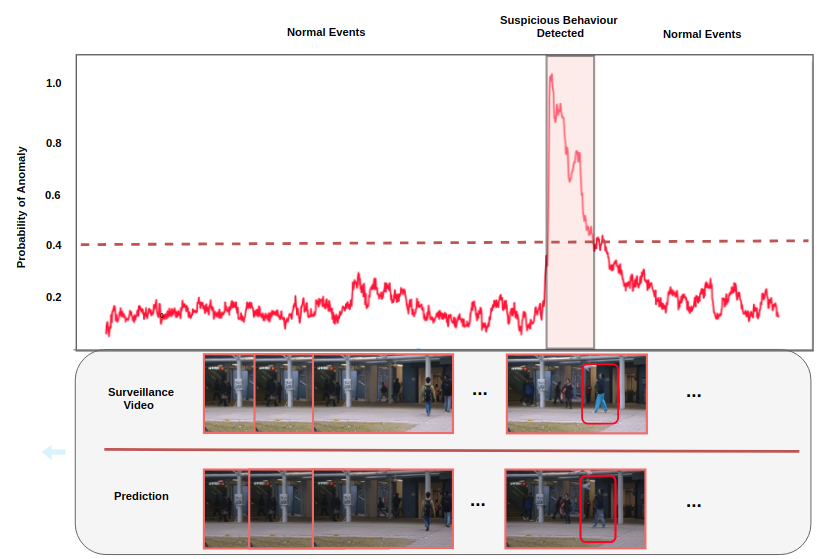}
    \caption{An example of our proposed video anomaly detection method. Our method uses a future frame prediction framework. Given several observed frames in a video, our model predicts the future frame. If the future frame is an anomaly, the predicted future frame is likely to be very different from the actual future frame. This reconstruction error allows us to detect the anomaly in a video.} 
    \label{fig:introduction}
\end{figure}

Some recent work~(e.g.~\cite{hasan2016learning,lu2013abnormal,luo2017remembering,sabokrou2018adversarially}) in anomaly detection uses the idea of frame reconstruction. They build models that learn to reconstruct the normal (or regular) frames observed during training. During testing, any irregular (abnormal) event will lead to a large reconstruction error. The higher reconstruction error indicates the possible abnormal event in the frame. Previous work~\cite{an2015variational,sabokrou2018adversarially,liu2018future} has applied variants of generative models such as variational autoencoder (VAE)~\cite{kingma2013auto} or generative adversarial network (GAN)~\cite{goodfellow2014generative} to model the distribution of the natural behaviours. To build a real-time anomaly detection system, Liu et al.~\cite{liu2018future} propose a future frame prediction framework for anomaly detection. Given several observed frames, their method learns a GAN-based model to predict the future frame. An anomaly then corresponds to a large difference between the predicted future frame and the actual future frame. One limitation of \cite{liu2018future} is that it directly concatenates the several observed frames as the input to the GAN model. As a result, the model does not directly represent the temporal information in a video. Although \cite{liu2018future} uses optical flow features which capture some temporal information at the feature level, the optical flow information is only used as a constraint during training and is not used during testing. 

In this paper, we follow the future frame prediction framework in \cite{liu2018future} and propose a new approach that better capture the temporal information in a video for anomaly detection. We propose to combine sequential models (in particular, ConvLSTM) with generative models (in particular, VAE) to build a model that can be trained end-to-end. Although sequential generative models have been previously proposed for speech recognition and music generation \cite{mogren2016c, chung2015recurrent}, they have not been applied in anomaly detection. An example of our proposed video anomaly detection system can be seen in Fig \ref{fig:introduction}. Given several consecutive frames, our model learns to predict the next future frame. For normal frames, our method is able to predict the next frame reasonably well. When there is anomaly in the future frame, the prediction is often distorted and blurry. By comparing the predicted future frame with the actual future frame, our system can detect suspicious behaviours or events (in this case, the man is throwing his bag up and down) are detected in a video frame. 

In this paper, we make the following contributions. We propose a sequential generative model for video anomaly detection using the future frame prediction framework.  We combine ConvLSTM with VAE to better capture the temporal relationship among frames in a video. Our experimental results demonstrate that the proposed model outperforms existing state-of-the-art approaches, even without using optical flow features. 

\section{Related Work}
\label{sec:relatedwork}
In this section, we review several lines of prior research related to our work.

\noindent{\bf Anomaly Detection with Hand-crafted Features}: Early work in video anomaly detection uses hand-crafted features. \cite{tung2011goal,wu2010chaotic} use trajectory features to represent normal behaviours. However, these methods can not be applied to crowded scenes. To address this limitation, low-level features such as histogram of oriented gradient and histogram of oriented flows are also applied \cite{dalal2005histograms,dalal2006human} for human detection. \cite{zhao2011online,lu2013abnormal,cong2011sparse} represent each scene by a dictionary of temporal and spatial information. These approaches have low performance due to the fact that the dictionary does not ensure the capacity of normal events and cannot classify anomaly correspondingly. Statistical-based models have also been proposed. For example, \cite{kim2009observe} proposes an approach based on a mixture of probabilistic PCA (MPPCA) with optical flow pattern. Gaussian mixture model \cite{mahadevan2010anomaly} has also been applied for anomaly detection.

\noindent{\bf Anomaly Detection with Deep Learning}: In order to address the limitation of hand-crafted features in anomaly detection, there has been recent work that explores the use of deep learning approaches. A lot of these methods learn a deep learning model to reconstruct a frame and use the reconstruction error for anomaly detection. Inspired by \cite{masci2011stacked}, Hasan et al.~\cite{hasan2016learning} apply convolutional autoencoder for reconstructing normal frames. Some follow-up works~\cite{sabokrou2016video,chalapathy2017robust} propose to build a more robust version. Xu et al.~\cite{xu2017detecting} use stacked de-noising autoencoders~\cite{vincent2008extracting} and optical flow to capture both appearance and motion information.

Some work considers using a future frame prediction approach for anomaly detection. Medel et al. \cite{medel2016anomaly} apply ConvLSTM as a backbone network and build a future prediction model for anomaly detection. Luo et al. \cite{luo2017remembering} combine autoencoder and ConvLSTM to reconstruct the output of ConvLSTM to the original image size. Because the inner structure of ConvLSTM is entirely deterministic, these predictive modeling methods cannot predict highly structured moving objects, which results in inaccurate predictions of anomalies.

Generative models, such as VAE~\cite{kingma2013auto} and GAN~\cite{goodfellow2014generative}, have been applied for the purpose of learning the distribution of regular frames. Sabokrou et al.~\cite{sabokrou2018adversarially} propose a one class classifier using conditional adversarial networks \cite{isola2017image}. Xie et al.~\cite{xie2012image} use a GAN-based image inpainting method to detect and localize the abnormal objects. Liu et al. \cite{liu2018future} propose a GAN-based future frame prediction network with optical flow network\cite{dosovitskiy2015flownet}.  An et al. \cite{an2015variational} apply VAE to build an anomaly detection system, but the method is not performed on real-world datasets. 

\noindent{\bf Sequential Generative Models}: There has been some work on incorporating sequential information in generative models. Chung et al.~\cite{chung2015recurrent} argue that latent random variables can play crucial roles in the dynamics of RNN. By combining VAE and RNN, they are able to model sequences with significant improvement on RNN. However, this model has only been used on simple tasks such as speech generation or handwriting generation. \cite{mogren2016c} propose a sequential generative model using adversarial training on RNN. They argue that with the supervision of a discriminator, their proposed generative model can be trained to be very expressive with high flexibility on continuous sequences such as music. However, the potential of this model on computer vision tasks has not yet been explored.

\section{Background}
\label{sec:background}
\begin{figure*}[t]
    \centering
    \includegraphics[width= 1.0\textwidth]{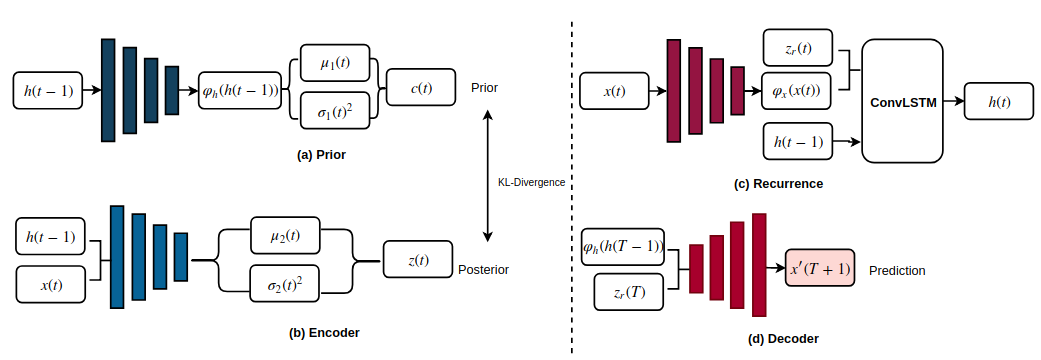}
    \caption{An overview of our proposed Conv-VRNN model at one time-step of a sequence. Our model requires 4 steps to process the input: (a) calculating the prior distribution in VAE; (b) encoder for posterior distribution and latent variable; (c) recurrence module for sequence modelling; (d) decoder for prediction.}
    \label{fig:VRNN}
\end{figure*}
\subsection{Variational Autoencoder}
Variational autoencoder (VAE) \cite{kingma2013auto} has been shown to be effective in reconstructing complex distributions for non-sequential data. Given an input $x$, VAE applies an encoder (also known as inference model) $q_\theta(z|x)$ to generate the latent variable $z$ that captures the variation in $x$. It uses a decoder $p_\phi(x|z)$ to approximate the observation given the latent variable. The inference model represents the approximate posterior using the mean $\mu$ and variance $\sigma^2$ calculated by a neural network $q_\theta(z|x) \sim \mathcal{N}(\mu_x,\,\sigma^2_x)$, where $\mu_x$ and $\sigma^2_x$ are outputs of some neural networks that take $x$ as the input. A prior $p(z)$ is chosen to be a simple Gaussian distribution. With the constraints of distribution on latent variables, the complete objective function can be described as below:
\begin{equation}
\begin{split}
L(x | \theta,\phi) = -KL(q_\theta(z|x)||p(z))  + \\ \mathbb{E}_{q_\theta(z|x)}[logp_\phi(x|z)]
\end{split}
\end{equation}
where $KL(q_\theta(z|x)||p(z))$ is the Kullback-Leibler divergence \cite{hershey2007approximating} between the prior and the posterior. 

\subsection{Variational Recurrent Neural Network} VAE is a generative model. It cannot directly be used to model sequential data. For the problem of anomaly detection, our data are inherently sequential since we need to consider the information in several consecutive frames in order to predict the next frame. Variational Recurrent Neural Network (VRNN)~\cite{chung2015recurrent} is an extension of vanilla VAE. It combines VAE with a recurrent neural network in order to model sequential data. Since this approach shares the same inspiration with our Conv-VRNN approach, we will explain the technical details in the next section.

\section{Approach}
\label{sec:approaches}

Following \cite{liu2018future}, we approach the anomaly detection problem using the future frame prediction framework. The goal is to build a model that takes several frames in a video as the input and predict the future frame. The predicted future frame is then compared with the actual future frame. If their difference is significant, we will consider it to be an anomaly. The main difference from \cite{liu2018future} is that our proposed approach combines a recurrent network with a generative model. As a result, our approach can better capture temporal information in the video.

Our problem formulation is as follows. Given a sequence of frames $x(1),...,x(T)$, we aim at predicting the next frame $x(T+1)$. Note that $T$ is a constant which we define as 4 in our case. We use $x'(T+1)$ to denote the predicted frame at time $T+1$. During training, we learn a model that minimizes the difference between the predicted and actual future frames, i.e. $L = |(x(T+1)-x'(T+1)|$. During testing, if this difference is too large, we will consider $x(T+1)$ to be an anomaly.

In this section, we first introduce our model Conv-VRNN (Sec.~\ref{sec:model1}) for future frame prediction. Our model combines VAE and a ConvLSTM module. We then describe how to use the proposed model to detect anomaly during testing (Sec.~\ref{sec:test}).

\subsection{Conv-VRNN for Future Frame Prediction}\label{sec:model1}
To extend VAE to model image sequences for anomaly detection, we use the idea of Variational Recurrent Neural Network (VRNN)~\cite{chung2015recurrent} and build a Conv-VRNN model for future frame prediction. An overview of our proposed model is shown in Figure \ref{fig:VRNN}. Let $x(t) \in \mathbb{R}^{H\times W \times 3}$ be the input image at time $t$, where $H\times W$ is the spatial dimension of the image. We define $h(t)\in \mathbb{R}^{H\times W \times 3}$ to be the hidden state of a ConvLSTM at time step $t$. Note that we choose the spatial dimension of $h(t)$ to match the image size. Our method consists of four components at each time step $t$:

\noindent\textbf{Prior Distribution in VAE:} This module takes the hidden state $h(t-1)$ from the previous time step as the input. It then generates a distribution on the latent variable in VAE. We first extract a feature vector from $h(t-1)$. Since $h(t-1)\in \mathbb{R}^{H\times W \times 3}$ is a 3D tensor and can be treated as a image, we can use a standard convolutional neural network (CNN) to extract the feature from $h(t-1)$. We denote this feature as $\varphi_h(h(t-1))\in\mathbb{R}^{H'\times W'\times F}$, where $H'\times W'$ and $F$ correspond to the spatial dimension and the channel dimension of the CNN feature map. Here we set $H'\times W'\times F=16\times16\times32$. We then apply two different fully connected layers on $\varphi_h(h(t-1))$ to produce two vectors corresponding to the mean and the variance of a Gaussian distribution in VAE, denoted by $\mu_{1}(t)$ and $\sigma_{1}(t)$. In our implementation, the dimension of $\mu_{1}(t)$ and $\sigma_{1}(t)$ is set to be 20, i.e. $\mu_{1}(t), \sigma_{1}(t)\in\mathbb{R}^{20}$. We then use $\mu_{1}(t)$ and $\sigma_{l}(t)$ to define a Gaussian distribution for the prior distribution on the latent variable in VAE as follows:
\begin{equation}
    c(t) \sim \mathcal{N}\left(\mu_{1}(t),\,diag\left(\sigma_{1}(t)^2\right)\right)
\label{eq:3}
\end{equation}
where $diag(\cdot)$ creates a diagonal matrix from a vector and $c(t)$ represent the prior distribution on the latent variable. 

\noindent\textbf{Encoder:} The module takes the hidden state $h(t-1)$ of previous time step $t-1$ and the frame $x(t)$ at current time $t$ as the input. It then produces a vector of the latent variable in VAE. We first concatenate $x(t)$ and $h(t-1)$ along their channel dimensions, then apply a CNN to extract a feature map. Again, we apply two different fully connected layers on this feature map to produce $\mu_2(t)$ and $\sigma_2(t)$. Similarly, the dimension of $\mu_2(t)$ and $\sigma_2(t)$ to be 20. We then define the posterior of the latent variable $z(t)$ in VAE as:
\begin{equation}
\begin{split}
    q_\theta\left(z(t)|concat\left(x(t),h(t-1)\right)\right) \\
    \sim \mathcal{N}\left(\mu_2(t),\,diag\left(\sigma_2(t)^2\right)\right)
\label{eq:4}
\end{split}
\end{equation}
where $z(t)\in\mathbb{R}^{20}$. To measure the distribution loss between Eq.~\ref{eq:3} and Eq.~\ref{eq:4} at time step $t$, we can use the KL-divergence metric $KL\left(q_\theta\left(z\left(t\right)|x\left(t\right),h\left(t-1\right)\right)||c\left(t\right)\right)$.

\noindent\textbf{Recurrence:} To capture the temporal information among frames in a video, we use a ConvLSTM to represent the recurrent relationship among frames. From the current input image $x(t)$, we apply a CNN to extract a feature map which we denote as  $\varphi_x(x(t))\in\mathbb{R}^{H'\times W'\times F}$. To match the dimension of this feature, we also resize the latent variable $z(t)$ (recall $z(t)\in\mathbb{R}^{20}$) as follows. We first use fully connected layers to map $z(t)$ to a high-dimensional space $\mathbb{R}^{1024}$, then reshape to a 3D tensor of dimension $H'\times W'\times F = 16\times16\times32$. We use $z_{r}(t)\in\mathbb{R}^{H'\times W'\times F}$ to denote this reshaped tensor. We concatenate the input feature $\varphi_x(x(t))$ with the  $z_r(t)$ along the channel dimension and use it as the input to ConvLSTM at time $t$:
\begin{equation}
\begin{split}
\hspace{-10pt} h\left(t\right) = f_{ConvLSTM}\left(concat\left(\varphi_x\left(x\left(t\right)\right),z_r\left(t\right)\right), h\left(t-1\right)\right)
\end{split}
\end{equation}

\noindent\textbf{Decoder:} This module takes the resized hidden state $z_r(t)$ as its input and produces a predicted frame $x'(t+1)$ for the next time-step. Note that the dimensions of $z_r(t)$ match those of the extracted feature of previous hidden state $\varphi_{h}(h(t-1))$.  We concatenate $z_r(t)$ and $\varphi_{h}(h(t-1))$ along the channel dimension. The result is used as the input of this decoder module. The decoder is implemented as a deconvolutional nerual network that generates the predicted frame $x'(t+1)\in\mathbb{R}^{H\times W\times 3}$. 


\noindent\textbf{Model Learning:} For learning parameters in Conv-VRNN, we combine the least absolute deviation ($L1$ loss) \cite{pollard1991asymptotics}, multi-scale structural similarity measurement (msssim loss) \cite{wang2003multiscale} and gradient difference (gdl loss) \cite{mathieu2015deep} to define a loss that measure the quality of the predicted frame. These three loss functions can be defined as follows:\\
(1) L1 loss between ground-truth and prediction is the summation of the absolute value between every pixel of the two images. \\
(2)We use multi-scale SSIM to represent the structural difference. MSSSIM is a multi-scale version of SSIM, which performs better on video sequences.  \\
(3) Gradient difference is widely used for measuring the performance of a prediction. Gradient difference loss considers the intensities difference between neighbour pixels. 

Overall, given the predicted frame $x'(T+1)$ and the ground-truth $x(T+1)$, the complete loss function is defined as:
\begin{equation}
\begin{split}
    L_{prediction} =  L_1(x(T+1),x'(T+1))
    \\+L_{msssim}(x(T+1),x'(T+1))
    \\+L_{gdl}(x'(T+1),x'(T+1))
\end{split}
\label{eq:7}
\end{equation}

We define the complete objective function as:
\begin{equation}
\begin{split}
    L = \sum_{t=1}^{T}(-KL( q_\theta(z(t)|x(t),h(t-1))||c(t))) + L_{prediction}.
\end{split}
\end{equation}  

\subsection{Anomaly Detection}
\label{sec:test}
Given an input sequence of frames $x(1), x(2), ..., x(T)$ during testing, we use our model to predict the next frame $x'(T+1)$ in the future. This predicted future frame $x'(T+1)$ is compared with the ground-truth future frame $x(T+1)$ by calculating $L_{prediction}$ (see Eq. \ref{eq:7}).  Same as \cite{liu2018future}, after calculating the overall spatial loss of each testing video, we normalize the losses to get a score $S(t)$ in the range of $[0,1]$ for each frame in the video by: 
\begin{equation}
    S(t) = \frac{L_{prediction}(t)-\min L_{prediction}}{\max L_{prediction}-\min L_{prediction}}.
\end{equation}
We then use $S(t)$ as the score indicating how likely a particular frame is an anomaly. 
\section{Experiments}
\label{sec:experiments}
In this section, we first discuss our experimental setup in Sec.~\ref{sec:setup}. Then we present both quantitative and qualitative results in Sec.~\ref{sec:results}. We also perform extensive ablation studies in Sec.~\ref{sec:ablation} to analyze our proposed approach.

\subsection{Experimental Setup}\label{sec:setup}
\begin{figure}
  \centering
  \begin{tabular}{cc}
    \includegraphics[width=3.5cm]{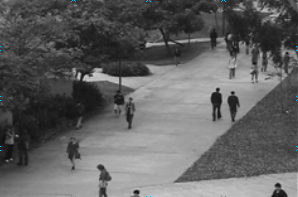}&
    \includegraphics[width=3.5cm]{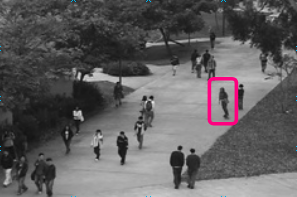}\\
    \includegraphics[width=3.5cm]{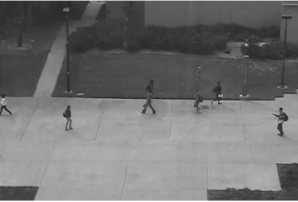}&
    \includegraphics[width=3.5cm]{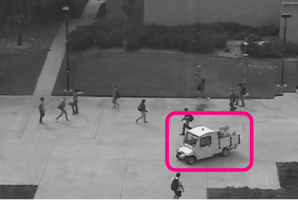}\\
    \includegraphics[width=3.5cm]{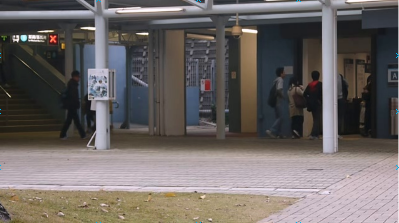}&
    \includegraphics[width=3.5cm]{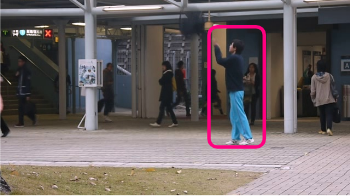}\\
    normal & abnormal\\
  \end{tabular}
  \caption{Example frames from the three datasets. 1st row: UCSD Pedestrian 1 (Ped1) dataset; 2nd row: UCSD Pedestrian 2 (Ped2) dataset; 3nd row: CUHK Avenue dataset. We show both normal and abnormal frames from these datasets. The abnormal behaviours are indicated by the red bounding box. Note that the red bounding box is only for visualization purpose and is not used during training.}
  \label{fig:dataset}
\end{figure}

\paragraph{Datasets:} We evaluate our method on three benchmark datasets. (1) UCSD Pedestrian 1 (Ped 1) dataset\cite{mahadevan2010anomaly}: this dataset contains 34 training videos and 36 testing videos. In training videos, only pedestrians exist in the frames. Test videos include 40 abnormal events, such as moving bicycles and vehicles. (2) UCSD Pedestrian 2 (Ped 2) dataset\cite{mahadevan2010anomaly}. This dataset considers the same set of anomalies with the UCSD Ped 1 dataset. It consists of 16 training videos and 12 testing videos with 12 irregular occasions. (3) CUHK Avenue (Avenue) dataset \cite{lu2013abnormal}. This dataset consists of 16 training videos and 21 testing videos. It contains 47 abnormal events like throwing things, wandering, and running. Figure~\ref{fig:dataset} shows some example frames from these datasets.

\paragraph{Evaluation Metric:} Following prior work \cite{liu2018future} \cite{luo2017revisit} \cite{mahadevan2010anomaly}, we evaluate our methods using the area under the ROC curve (AUC). The ROC curve is obtained by varying the threshold for the anomaly score. A higher AUC value represents a more accurate anomaly detection system. To ensure the comparability between different methods, we calculate AUC from the frame-level prediction, which has been used by different existing methods. 







\subsection{Experimental Results}\label{sec:results}

\begin{table}[]
\centering
\caption{Comparison of different methods in terms of AUC on UCSD Ped1, UCSD Ped2 and CUHK Avenue datasets.}
\begin{tabular}{|l|l|l|l|}
\hline
 & Ped1 & Ped2& Avenue \\ \hline
MPCCA \cite{kim2009observe} & 59.0\% & 69.3\% & N/A \\ \hline
Del et al.\cite{del2016discriminative} & N/A & N/A & 78.3\% \\ \hline
Conv-AE \cite{hasan2016learning} & 75.0\% & 85.0\% & 80.0\% \\ \hline
ConvLSTM-AE \cite{luo2017remembering}& 75.5\% & 88.1\% & 77.0\% \\ \hline
Stacked RNN \cite{luo2017revisit} & N/A & 92.2\% & 81.7\% \\ \hline
Liu et al. \cite{liu2018future} & 83.1\% & 95.4\% & 84.9\% \\ \hline
\textbf{Conv-VRNN (ours)} & \textbf{86.26\%} & \textbf{96.06\%} & \textbf{85.78\%} \\  \hline
\end{tabular}
\label{table:1}
\end{table}

\begin{figure*}
  \centering
  \begin{tabular}{cccc}
    \includegraphics[width=3.5cm]{Ped1-GT_normal.png}&
    \includegraphics[width=3.5cm]{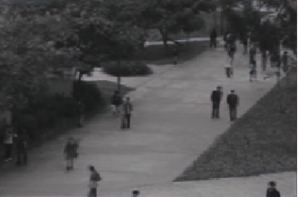}&
    \includegraphics[width=3.5cm]{Ped1-GT_abnormal.png}&
    \includegraphics[width=3.5cm]{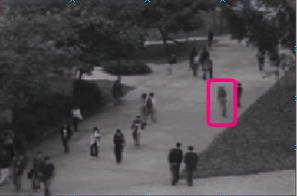}\\
    \includegraphics[width=3.5cm]{Ped2-GT_normal.png}&
    \includegraphics[width=3.5cm]{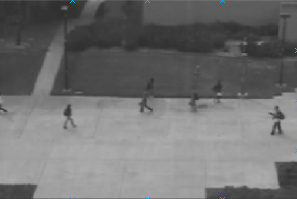}&
    \includegraphics[width=3.5cm]{Ped2-GT_abnormal.png}&
    \includegraphics[width=3.5cm]{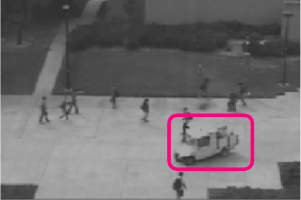}\\
    \includegraphics[width=3.5cm]{Avenue-GT_normal.png}&
    \includegraphics[width=3.5cm]{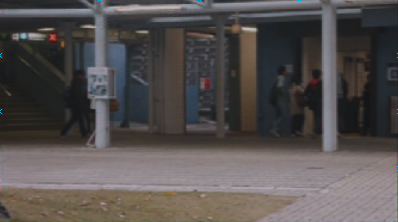}&
    \includegraphics[width=3.5cm]{Avenue-GT_abnormal.png}&
    \includegraphics[width=3.5cm]{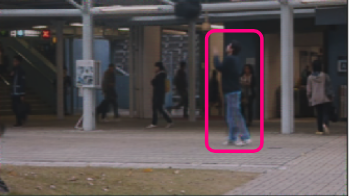}\\
    GT (normal) & prediction (normal) & GT (abnormal) & prediction (abnormal)\\
  \end{tabular}
  \caption{Examples of frame predictions on three datasets. The 1st row shows predicted frames that are normal. The 2nd row shows predicted frames with anomalies. For an abnormal frame, the predicted frame tends to be blurry and distorted. The bounding boxes are for visualization purpose and are not part of the model prediction.}
  \label{fig:result_frame}
\end{figure*}

\begin{figure*}
  \centering
  \begin{tabular}{cc}
  
    \includegraphics[height=5cm, width= 8cm]{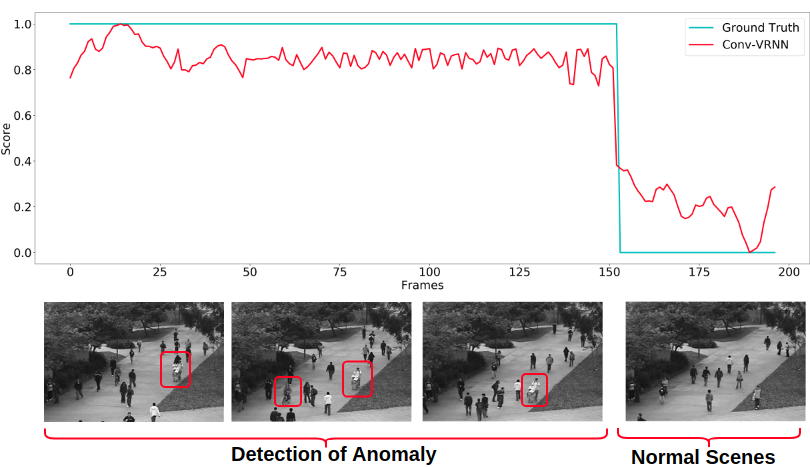}&
    \includegraphics[height=5cm, width= 8cm]{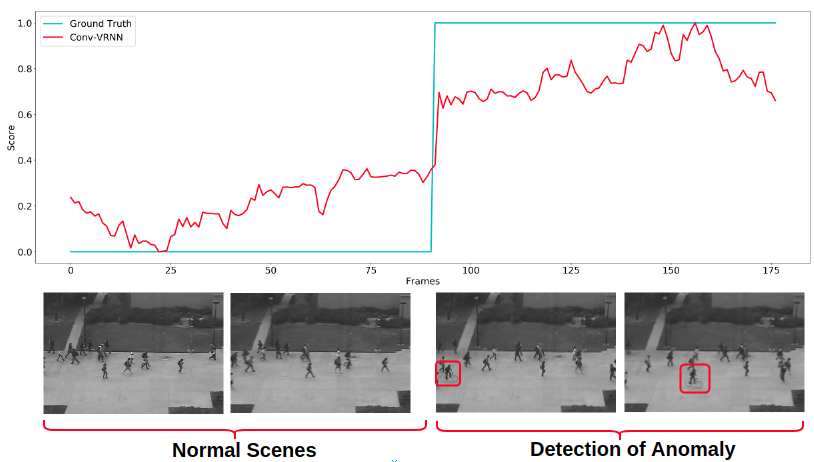}\\
    Ped1&Ped2\\
  \end{tabular}
  \begin{tabular}{c}
    \includegraphics*[height=7cm, width= 17cm]{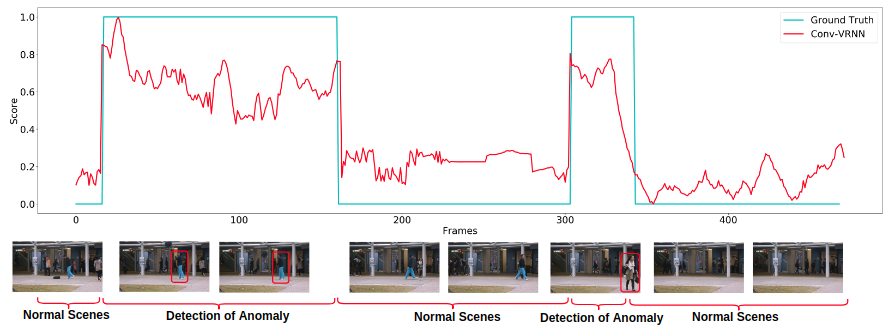}\\
    Avenue\\
  \end{tabular}
  \caption{Examples of anomaly detection on three datasets. We plot the anomaly score of our model and the ground-truth anomaly score. Again, the bounding boxes are for visualization purpose.}
  \label{fig:result_ped}
\end{figure*}

Table~\ref{table:1} shows the results of our proposed method compared with existing state-of-the-art approaches. To be consistent with \cite{liu2018future}, we have set $T=4$. In other words, our model takes 4 consecutive frames as the input and predicts the future frame at the next time step. It then compares the prediction with the actual frame at the next time step to decide whether this frame is an anomaly. We can see that Conv-VRNN outperforms existing methods on all three datasets.

Figure~\ref{fig:result_frame} shows some qualitative examples of future frame prediction. We can see that for a normal frame, the predicted future frame tends to be close to the actual future prediction. For an abnormal frame, the predicted future frame tends to be blurry or distorted compared with the actual future frame. Figure~\ref{fig:result_ped} shows example of detected anomaly by visualizing the anomaly score on different frames in a video. 
 


\subsection{Ablation Study}\label{sec:ablation}
We perform additional ablation study to gain further insights of our proposed methods. 

\subsubsection{Conv-VAE vs Conv-VRNN}
In order to analyze the effect of incorporating temporal information, we implement a variant of our model without RNN. We call this variant Conv-VAE. Conv-VAE uses the encoder module to encode a latent variable and uses the decoder module for prediction. We have experimented with Conv-VAE that takes either one input frame or four frames to predict the next frame. The results are shown in Table~\ref{tab:2}. We can see that Conv-VRNN outperforms Conv-VAE. This demonstrates the importance of capturing the temporal information using RNN for anomaly detection.  

\begin{table}[]
\centering
\caption{Comparision of Conv-VAEs versus Conv-VRNN in terms of AUC on three datasets.}
\begin{tabular}{|l|l|l|l|}
\hline
                         & Ped 1           & Ped 2           & Avenue          \\ \hline
Conv-VAE & 82.42\%          & 89.18\%          & 81.82\%          \\ \hline
Conv-VRNN & \textbf{86.26\%} & \textbf{96.06\%} & \textbf{85.48\%} \\ \hline
\end{tabular}
\label{tab:2}
\end{table}

\subsubsection{Analysis on Losses}
As we mentioned in Sec \ref{sec:approaches}, we apply three different losses for prediction. The analysis of the impact of the losses can be visualized in Table \ref{tab:loss}. We choose three combinations of objective functions for evaluation: constraint only on intensity ($L_{1}$), constraint on intensity and structure ($L_1+L_{msssim}$), constraint on intensity, structure and gradient ($L_1+L_{msssim}+L_{gdl}$). The results demonstrate that the appearance information is better captured by the model with more constraints.

\begin{table}[]
\centering
\caption{Evaluation of different combinations of various loss terms in the objective functions in our Conv-VRNN network on the Ped1 dataset. The results show that the combination of all loss terms gives the best performance.}
\label{tab:loss}
\begin{tabular}{|c|c|c|c|}
\hline
\textbf{$L_{1}$} & \cmark & \cmark & \cmark \\ \hline
$L_{msssim}$ & \xmark & \cmark & \cmark \\ \hline
$L_{gdl}$ & \xmark & \xmark & \cmark \\ \hline
$AUC$ & 80.29\% & 83.34\% & \textbf{86.26\%} \\ \hline
\end{tabular}%
\end{table}

\begin{figure*}
    \centering
    \begin{tabular}{ccc}
      \includegraphics[height=3.5cm]{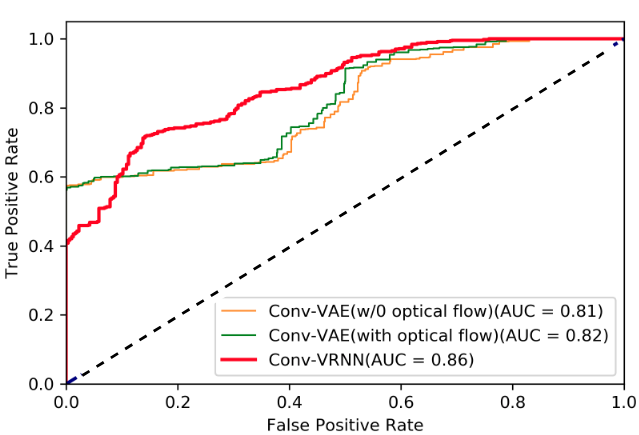}&
      \includegraphics[height=3.5cm]{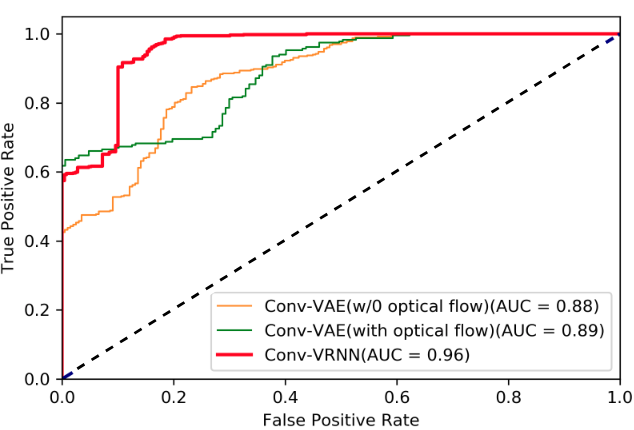}&
      \includegraphics[height=3.5cm]{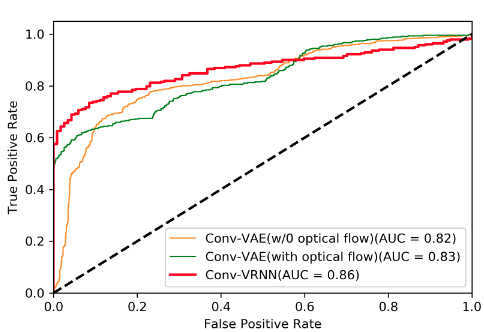}\\
      Ped1 & Ped2 & Avenue\\
    \end{tabular}
    \caption{ROC curves of our Conv-VRNN method, Conv-VAE~(w/o optical flow) and Conv-VAE~(with optical flow) on three datasets.}
    \label{fig:ROC1}
\end{figure*}
\subsubsection{Sequential Model vs Optical Flow} Our Conv-VRNN uses a RNN module to capture the temporal information in a video. An alternative way of capturing temporal information is to use optical flow features. We have implemented a Conv-VAE model with such constraint. Following \cite{liu2018future}, we apply the pretrained Flownet~\cite{dosovitskiy2015flownet} to estimate the optical flow, and use the returned loss of the Flownet as a motion constraint of the network only in training time. Table \ref{tab:3}, Figure~\ref{fig:ROC1} show that although adding optical flow in our implementation of Conv-VAE improves the performance compared with Conv-VAE applied on only raw frames, our proposed Conv-VRNN approach still performs better even if we do not use optical flow features. This demonstrates that it is more effective to design the generative model to directly capture the temporal information instead of relying on low-level optical flow features. 
\begin{table}[]
\centering
\caption{Comparison between our Conv-VRNN model with different VAE-based models (with or without optical flow features). Our proposed Conv-VRNN outperforms Conv-VAE (with optical flow) even if our model does not use optical flow features.}
\begin{tabular}{|l|l|l|l|}
\hline
 & Ped1 & Ped2&Avenue \\ \hline
Conv-VAE(w/o optical flow) & 80.15\% & 88.13\% & 80.92\% \\ \hline Conv-VAE(with optical flow) & 81.36\% & 89.52\% & 82.23\%\\ \hline
Conv-VRNN & \textbf{86.26\%} & \textbf{96.06\%} & \textbf{85.78\%}\\ \hline
\end{tabular}
\label{tab:3}
\end{table}

\section{Conclusion}
\label{sec:conclusion}
In this paper, we have proposed a sequential generative network for anomaly detection based on convolutional VRNN using the future frame prediction framework. By combining a ConvLSTM module with VAE, our approach can effectively capture the temporal information crucial for future frame prediction. On three benchmark datasets, our proposed approach outperforms existing state-of-the-art methods.

\noindent {\bf Acknowledgement}: This work was supported by the NSERC and UMGF funding programs. We thank NVIDIA for donating some of the GPUs used in this work.

{\small
\bibliographystyle{ieee}
\bibliography{egbib.bib}
}

\end{document}